%% file: arxiv.tex
\documentclass{article}

\usepackage[preprint]{neurips_2024}

\usepackage[utf8]{inputenc} 
\usepackage[T1]{fontenc}    
\usepackage{hyperref}       
\usepackage{url}            
\usepackage{booktabs}       
\usepackage{amsfonts}       
\usepackage{nicefrac}       
\usepackage{microtype}      
\usepackage{xcolor}         

\newcommand{\fxh}[1]{\textcolor{blue}{[#1--fxh]}}

\usepackage{amsmath}
\usepackage{graphicx}
\usepackage{subcaption}
\usepackage{amsthm}
\usepackage{amssymb}
\usepackage{multirow}
\usepackage{wrapfig,lipsum}
\usepackage{algorithm}
\usepackage{enumitem}
\usepackage{algpseudocode}
\usepackage{mathrsfs}
\usepackage{ulem}
\usepackage{comment}
\usepackage{tabularx}
\usepackage{booktabs}
\usepackage{subcaption}

\makeatletter
\newcommand{\rmnum}[1]{\romannumeral #1}
\newcommand{\Rmnum}[1]{\expandafter\@slowromancap\romannumeral #1@}
\makeatother

\title{Integration of Large Language Models and \\ Federated Learning}

\author{%
  Chaochao Chen \\
  Zhejiang University\\
  \texttt{zjuccc@zju.edu.cn} \\
  \And
  Xiaohua Feng \\
  Zhejiang University\\
  \texttt{fengxiaohua@zju.edu.cn} \\
  \And
  Yuyuan Li \\
  Hangzhou Dianzi University\\
  \texttt{y2li@zju.edu.cn} \\
  \AND
  Lingjuan Lyu \\
  Sony AI\\
  \texttt{lingjuan.lv@sony.com} \\
  \And
  Jun Zhou \\
  Ant Group\\
  \texttt{junzhougucas@gmail.com} \\
  \And
  Xiaolin Zheng \thanks{Corresponding authors.} \\
  Zhejiang University\\
  \texttt{xlzheng@zju.edu.cn} \\
  \AND
  Jianwei Yin $^*$ \\
  Zhejiang University\\
  \texttt{zjuyjw@zju.edu.cn} \\
}

\begin{document}

\maketitle

\begin{abstract}
As the parameter size of Large Language Models (LLMs) continues to expand, there is an urgent need to address the scarcity of high-quality data. 
In response, existing research has attempted to make a breakthrough by incorporating Federated Learning (FL) into LLMs. 
Conversely, considering the outstanding performance of LLMs in task generalization, researchers have also tried applying LLMs within FL to tackle challenges in relevant domains. 
The complementarity between LLMs and FL has already ignited widespread research interest. In this paper, we aim to deeply explore the integration of LLMs and FL. 
We propose a research framework, dividing the fusion of LLMs and FL into three parts: the combination of LLM sub-technologies with FL, the integration of FL sub-technologies with LLMs, and the overall merger of LLMs and FL. 
We first provide a comprehensive review of the current state of research in the domain of LLMs combined with FL, including their typical applications, integration advantages, challenges faced, and future directions for resolution. 
Subsequently, we discuss the practical applications of the combination of LLMs and FL in critical scenarios such as healthcare, finance, and education, and provide new perspectives and insights into future research directions for LLMs and FL.
\end{abstract}

\input{body/1-intro}
\input{body/2-background}
\input{body/3-1-analysis}
\input{body/3-2-analysis}
\input{body/3-3-analysis}
\input{body/4-discussion}

\input{body/5-conclusion}

\section*{Acknowledgment}
This work was supported by National Key R\&D Program of China 2022YFB4501500, the Fundamental Research Funds for the Central Universities 226-2024-00241, and Ant Group. We thank all team members and partners involved in this study for their support and contributions. Additionally, we appreciate the valuable comments and suggestions provided by the reviewers of this paper.

\small
\bibliographystyle{plain}


\end{document}

%% file: body/1-intro.tex
\section*{1. Introduction} \label{sec-intro}

The advent of Large Language Models~\cite{wei2022emergent} (LLMs) has markedly influenced contemporary society.
These models use deep learning strategies, principally the transformer architecture~\cite{luitse2021great} to discern intricate patterns and structures inherent to data~\cite{adnan2019analytical}. 
Presently, a vast amount of work~\cite{raffel2020exploring,brown2020language,ouyang2022training} confirms that these models exhibit superior performance both in predefined tasks and practical applications. 
Impressively, given accurate instructions and demonstrations, these models are capable of adapting to specific contexts or addressing new tasks without additional fine-tuning, as corroborated by numerous studies~\cite{zhou2023least,kojima2022large,victor2022multitask}.
Moreover, LLMs have made significant strides in specialized domains, delivering commendable outcomes in areas like healthcare~\cite{wang2023huatuo}, finance~\cite{Cornucopia-LLaMA-Fin-Chinese}, law~\cite{huang2023lawyer, nguyen2023brief, dai2023laiw}, scientific knowledge analysis~\cite{taylor2022galactica}, and code generation~\cite{nijkamp2022codegen, li2023starcoder}.

\begin{figure}[htbp]%
\centering
\includegraphics[width=0.9\textwidth]{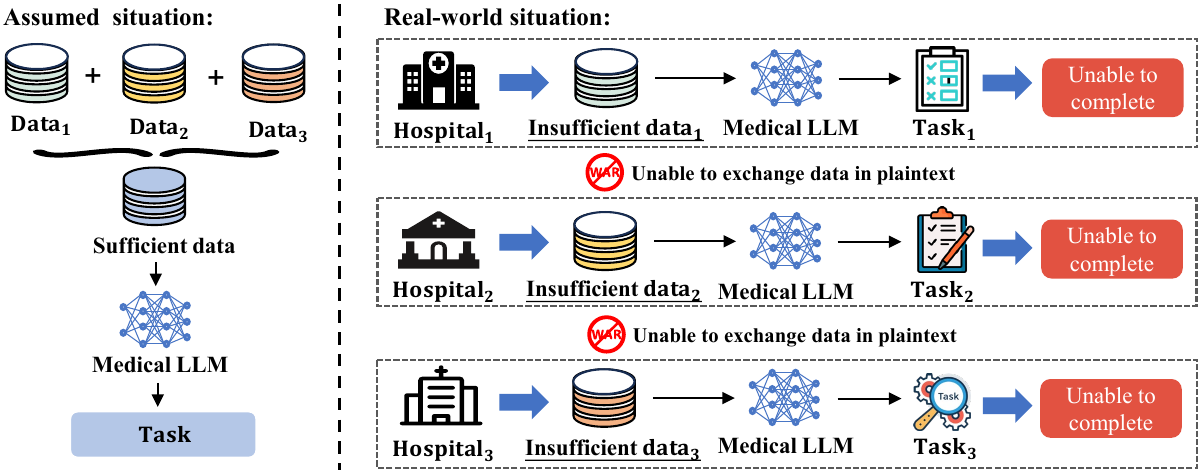}
\caption{The diagram illustrates the problem of data scarcity in LLMs. None
of the hospitals have enough data for training LLMs and they are reluctant to share data with each other.}\label{fig:problem}
\end{figure}

As the size of these models grows, more extensive training data is needed~\cite{kaplan2020scaling,hoffmann2022training}. 
However, recent research~\cite{villalobos2022will} points out that there is a gap between the slow growth of public domain data and the rapid expansion of training data needs. 
This discrepancy may result in a shortage of high-quality public domain data for LLM training.
Conversely, while private domains harbor colossal data volumes, concerns about privacy and commercial competition often hinder open collaboration and knowledge exchange.
Take Fig.~\ref{fig:problem} for example, suppose that three hospitals want to establish an LLMs for the medical field, their own datasets would likely be insufficient.
A joint dataset, on the other hand, would yield a substantial corpus. 
However, real-world data privacy regulations~\cite{albrecht2016gdpr} prevent direct plain text sharing between separate entities.

Considering the large parameter size and complex model structure of LLMs, common privacy-preserving computation techniques, such as Secure Multi-party Computation~\cite{cramer2015secure} (SMPC), Differential Privacy~\cite{dwork2014algorithmic} (DP), and Trusted Execution Environments~\cite{sabt2015trusted} (TEE), struggle to juggle privacy protection and computational efficiency effectively. 
Unlike these methods, Federated Learning~\cite{mcmahan2017communication} (FL) offers a more practical approach by allowing collaborative model development. 
FL demonstrates a mature engineering execution method and strikes an ideal balance between efficiency and data privacy~\cite{brauneck2023federated}. 
Therefore, a feasible solution to address the challenges of LLMs in practical applications is to introduce FL into the LLMs. 
Conversely, capitalizing on the strong task generalization capabilities of LLMs, they can also be employed within FL systems to help address challenges inherent to FL. 
Based on this complementarity, the combination of LLMs with FL has demonstrated exceptional performance benefits and mutual enhancement, a characteristic that has elicited widespread research interest.

In this paper, we focus on the promising direction of combining LLMs and FL.
Previous studies present initial perspectives on this integration~\cite{chen2023federated,zhuang2023foundation,yu2023federated}, providing preliminary insights into its motivations and future directions. 
Despite this, current research has not fully covered all areas related to the integration of LLMs and FL. 
Specifically, some studies focus on exploring the integration of sub-technologies within LLMs and FL~\cite{zhuang2023foundation}, neglecting the importance of the overall concept of Federated Large Language Models (FedLLMs). 
In view of this, we adopt a more comprehensive research approach to organize existing work on combining LLMs and FL.
%
%
By analyzing the current progress in research combining LLMs and FL, we offer unique insights into the benefits, challenges, and future development trends of their integration. 
Notably, while analyzing the combination of sub-technologies in LLMs and FL, we also explored sub-technologies shared by foundational models, extending beyond just language models to include multimodal and visual models. Since these shared sub-technologies can be easily adapted to language models, this broader perspective offers valuable insights into the integration of LLMs with FL.

The remainder of this paper is organized as follows: we first briefly introduce the technical backgrounds of FL and LLMs in Section 2. 
In Section 3, we present a comprehensive analysis of the current status, challenges, and future directions regarding the combination of LLMs and FL.
This includes three sub-sections: i) the integration of sub-technologies in LLMs with FL, ii) the integration of sub-technologies in FL with LLMs, and iii) the overall integration of LLMs and FL. 
Section 4 analyzes the application scenarios where LLMs are combined with FL.
Finally, in Section 5, we summarize the progress of research on the integration of LLMs and FL and present insights into the future development of this field.

%% file: body/2-background.tex
\section*{2. Background} \label{sec-back}

\subsection*{2.1 Large Language Models}
Language Models (LMs) aim to predict the probability distribution of future tokens based on a given sequence of tokens~\cite{sun2024trustllm}. As the size of model parameters and the amount of training data increase, LLMs have shown impressive capabilities in handling complex tasks, including In-context Learning (ICL)~\cite{brown2020language}, instruction following~\cite{victor2022multitask,ouyang2022training,wei2022finetuned}, and step-by-step reasoning~\cite{wei2022chain}. 
%

The success of LLMs is not just due to their larger model sizes and extensive training data but also owes much to the Transformer architecture~\cite{vaswani2017attention}.
Existing LLMs primarily rely on two design architectures~\cite{zhao2023survey}: only-decoder, and encoder-decoder~\cite{vaswani2017attention}, with the only-decoder architectures further divided into causal decoder~\cite{radford2019language,brown2020language} and prefix decoder~\cite{zhang2022examining}. 
Causal decoder architectures, which combine a unidirectional attention mask to ensure each input token can only attend to past tokens and itself~\cite{radford2018improving}, have been widely adopted across various existing LLMs, offering significant advantages with massive training data. 
Specifically, GPT-3~\cite{brown2020language} successfully demonstrated the effectiveness of this architecture.

Zhao et. al.~\cite{zhao2023survey} outline three key stages of training LLMs: pre-training, instruction-tuning, and alignment-tuning. 
During the pre-training stage, LLMs learn basic language processing abilities and world knowledge across a broad corpus, such as grammar, syntax, and general knowledge. 
%
%
%
Instruction-tuning becomes crucial for refining LLMs' ability to handle new tasks effectively.
It involves crafting precise task instructions or contextual learning strategies to bolster the model's adaptability to unseen tasks~\cite{wei2022finetuned}. 
Despite the benefits, there's a risk of instruction-fine-tuned models generating harmful content due to potential misalignment with human values~\cite{bender2021dangers,weidinger2021ethical}. 
Therefore, aligning LLMs with human values, such as honesty and harmlessness, through alignment-tuning has become an important task. 
To this end, InstructGPT~\cite{ouyang2022training} proposes alignment training methods, including supervised fine-tuning and reinforcement learning from human feedback~\cite{ouyang2022training}.

\subsection*{2.2 Federated Learning}

The concept of FL emerged to execute collaborative model learning with the data from participants while safeguarding their privacy~\cite{mcmahan2017communication}. 
Within FL, client devices asynchronously share updates such as weights and gradients while keeping raw data locally.
The Federated Averaging (FedAvg) algorithm~\cite{mcmahan2017communication}, which aggregates model updates from participating clients by averaging, is among the most prevalent aggregation algorithms in FL. 
%
%
%
Furthermore, studies~\cite{yang2019federated} consider the statistical challenges posed by heterogeneous user data in real-world scenarios where direct collaboration of data may be of poor quality, incomplete, or insufficient. 
In summary, these advancements greatly accelerate the development of FL, striking a balance between maintaining data quality and enhancing the efficiency of collaborative model creation.


%% file: body/3-1-analysis.tex
\section*{3. Analysis of Current Integration of LLMs and FL} \label{sec-analysis}


\begin{figure}[htbp]%
\centering
\includegraphics[width=1.0\textwidth]{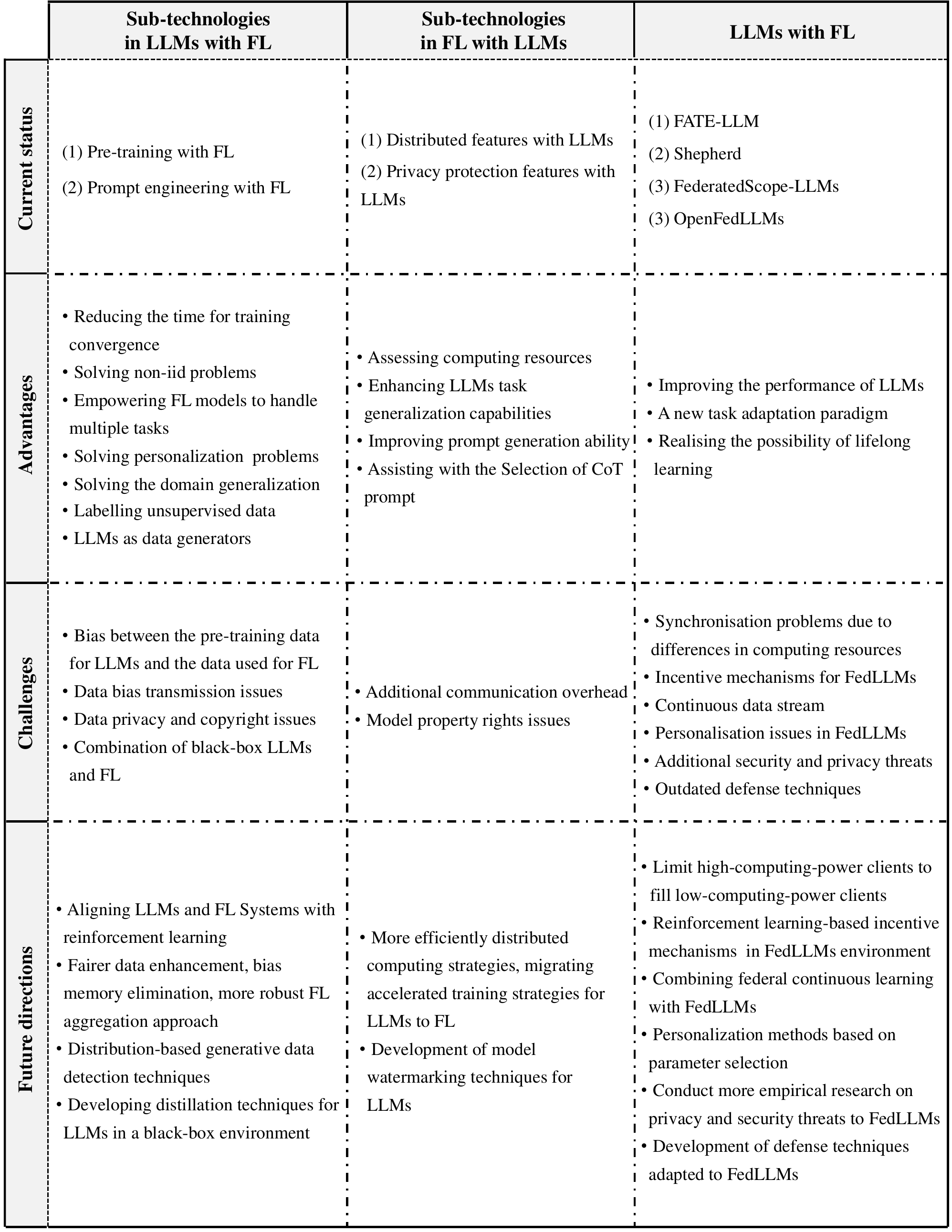}
\caption{Overview of the analysis process combining LLMs and FL. We sequentially analyze the integration of sub-technologies within LLMs with FL, the integration of sub-technologies within FL with LLMs, and the overall framework combining LLMs and FL. This includes the current status of integration, the advantages brought by the combination, potential challenges, and future directions for solutions.}\label{fig:framework}
\end{figure}

\subsection*{3.1 Integration of Sub-technologies in LLMs with FL}

Although FL has been widely applied in practice, unsolved issues remain. 
LLMs introduce novel solutions to FL by leveraging their pre-trained knowledge and generalization abilities for universal tasks.
Combining sub-technologies from LLMs with FL is a current focus of research exploration (Tabel \ref{tab:llms2fl}). 
Below, we analyze the research status of integrating sub-technologies from LLMs with FL, the challenges present, and possible future solutions.

\begin{table}[t]
    \caption{Overview of the current state of sub-technologies within LLMs and their integration with FL. We list the existing research on the combination of each sub-technology with FL and analyze the benefits they bring. Subsequently, we provide a brief summary of their methodologies.}
    \label{tab:llms2fl}
    \centering
    \small
    \resizebox{\linewidth}{!}{%
    \begin{tabular}{c|ccc}
    \toprule
    Sub-technologies & Advantages of integration & Summary of method & References \\
    \midrule
    \multirow{3}{*}{Pre-training} &
    Reducing the time for training convergence & Pre-trained models serve as the starting point for FL training & \cite{tan2022federated,liu2024language,nguyen2022begin} \\
     & Solving non-iid problems in FL & The server performs pre-training before commencing FL training & \cite{nguyen2022begin,chen2022importance} \\
     & Empowering FL models to handle multiple tasks & Modular design assigns each task to the corresponding module & \cite{agarwal2023practical,zhang2023next,zhang2023fedyolo} \\
    \midrule
    \multirow{2}{*}{Prompts engineering} &
    Solving personalization problems in FL & Personalized prompts are utilized to represent the local data distribution of client & \cite{yang2023efficient,li2023visual} \\
     & Solving the domain generalization problem in FL & Adaptive prompts for domain generalization are learned in a distributed manner & \cite{wei2023dual,bai2024diprompt} \\
    \bottomrule
    \end{tabular}
    }
\end{table}

\subsubsection*{3.1.1 Current Status}

Combining sub-technologies within LLMs with FL, current research primarily explores the following two aspects: pre-training and prompt engineering. Below, we detail the research status of each integration approach.

\paragraph{Pre-training.} 

Current research leverages pre-training techniques within LLMs to tackle multiple challenges in FL, such as reducing the time to convergence for FL training, addressing non-independent identically distributed (non-iid) issues in FL, and empowering FL models with the ability to handle multiple tasks.

\textit{Reducing the time for training convergence.} 
In FL research, starting with randomly initialized neural network weights often slows down model convergence. 
Studies show that using pre-trained models, trained on large datasets, as a starting point for FL can significantly reduce training time.~\cite{tan2022federated,liu2024language}. 
Instead of starting from scratch, clients can fine-tune the FL model using their local data.
Experiments demonstrate that using a pre-trained model can reduce the training time required to achieve a target error rate than those starting from random initialization~\cite{nguyen2022begin}. 
This faster convergence results in better-performing models in fewer communication rounds.

\textit{Solving non-iid problems in FL.} One common challenge in FL is dealing with data and system heterogeneity. 
Data heterogeneity refers to variations in data distributions across clients~\cite{kairouz2021advances}, while system heterogeneity relates to differences in client device capabilities~\cite{nguyen2022begin}.
To mitigate these challenges, researchers have developed joint optimization methods~\cite{huang2023generalizable,huang2023rethinking}. Starting FL from a pre-trained model initialization has been found to help alleviate the effects of data and system heterogeneity~\cite{nguyen2022begin}. This approach can lead to more stable global model aggregation and reduce the accuracy gap between FL and centralized learning, especially in scenarios with non-iid client data~\cite{chen2022importance}.

\textit{Empowering FL models to handle multiple tasks.} 
FL typically focuses on a single task, which may not be sufficient for real-world applications with diverse task requirements~\cite{zhuang2023mas}. Large pre-trained models have demonstrated the ability to perform well across multiple tasks~\cite{brown2020language}. Some research endeavors to integrate pre-trained models into the FL framework to enable FL models to handle various tasks~\cite{agarwal2023practical,zhang2023next}. However, more attention is needed on FL in mobile and edge devices. The FedYolo framework proposes a modular approach where clients load a complete pre-trained model and make future updates through communication-efficient modules~\cite{zhang2023fedyolo}. Experiments show that this design allows clients to simultaneously solve multiple unrelated tasks with a single pre-trained model, reducing catastrophic forgetting compared to full updates.

\paragraph{Prompts engineering.}

Prompt techniques have demonstrated exceptional performance within LLMs~\cite{guo2023promptfl}. 
Current research is exploring the integration of prompts with the FL framework to address issues of personalization, and domain generalization within FL.

\textit{Solving personalization problems in FL.} 
Personalized FL allows for personalized models to enhance their generalization and robustness by leveraging knowledge from distributed clients. 
The pFedPG framework has utilized large-scale pre-trained models to acquire robust representations while achieving efficient model personalization for heterogeneous clients~\cite{yang2023efficient}. 
While pFedPG does not consider client data characteristics, recent work, i.e., pFedPT,  uses personalized prompts to implicitly represent local data distributions~\cite{li2023visual}. During pFedPT training, each client generates a personalized prompt related to their data distribution, aiding classification tasks by incorporating this information into the aggregated model.

\textit{Solving the domain generalization problem in FL.} 
FL is crucial for learning from decentralized data, but faces challenges when training data (source domain) differs from the test dataset (target domain). The Fed-DPT framework initially addressed this using visual and textual prompts, but required domain labels during training and had limitations on the number of domains~\cite{wei2023dual}. To overcome this, the DiPrompT framework was proposed, learning adaptive prompts for domain generalization in a distributed manner~\cite{bai2024diprompt}. DiPrompT uses global prompts to capture shared knowledge and domain prompts for specific domain knowledge, eliminating the need for a strict one-to-one mapping between source domains and local clients.

\subsubsection*{3.1.2 Challenges and Future Directions}

The integration of sub-technologies within LLMs with FL can resolve many issues but also introduce some new challenges. Below, we discuss each of these new challenges.


\paragraph{Bias between the pre-training data for LLMs and the data used for FL.} 
The domain mismatch between training and test data poses a significant challenge in current research~\cite{glorot2011domain,long2015learning,tan2024heterogeneity}, especially when integrating sub-technologies from LLMs into FL. This mismatch can reduce the effectiveness of model transfer and application. Additionally, if synthetic data created by LLMs does not align with client data distribution, it may introduce bias and noise into the FL process.
To address these challenges, future research should focus on enhancing the quality and diversity of synthetic data generated by LLMs to closely match the underlying data distribution and application domains in FL. 
One potential approach is to utilize pre-processing techniques to fine-tune the alignment between LLMs and FL systems before incorporating them into FL processes~\cite{kaelbling1996reinforcement}. This strategy aims to minimize the bias between LLM-generated data and foundational FL data, ensuring their distributions are as similar as possible.

\paragraph{Data bias transmission issues.} 
In the era of LLMs, the training and fine-tuning datasets for LLMs are vast and diverse, potentially containing toxic content, user privacy data, politically sensitive information, and biases~\cite{nadeem2020stereoset}. LLMs, being probabilistic generative models with limited interpretability and controllability~\cite{singh2023augmenting}, may generate synthetic data of questionable quality and safety, leading to issues like data toxicity, biases, and misinformation. 
When FL training is conducted on these synthetic datasets, these problems can transfer to the final FL model.
To address these challenges, future research should focus on integrating LLMs into FL systems in a way that prevents new biases and avoids amplifying existing ones. This could involve developing LLMs data augmentation techniques guided by fairness principles and applying bias elimination techniques to remove biases from FL systems, such as combining LLM-based data augmentation with federated unlearning techniques. Additionally, creating more robust FL aggregation algorithms could effectively prevent the introduction of biases into the system.

\paragraph{Data privacy and copyright issues.} 
When LLM-generated data is used in FL, concerns about privacy rights and copyright emerge~\cite{chu2024protect}.
LLMs gather vast amounts of internet data during pre-training, including private and copyrighted information, making it hard to trace the origins of this data. Recent studies show that LLMs have strong memory capabilities~\cite{carlini2021extracting,carlini2023quantifying}, suggesting that the data they generate could closely resemble the privacy and copyright information encountered during training~\cite{li2024digger}. This poses legal risks for FL models trained using these datasets.
To address these issues, future research should explore how to balance the usefulness of synthetic data from LLMs with privacy and copyright protection. 
Firstly, it is imperative to develop a method for determining whether generated data adheres to privacy protection and copyright regulations. 
Building on this, researchers should be able to selectively generate data, ensuring distinct differentiation from the original data. 
Furthermore, exploring the interpretability of model inference within an FL environment is also a viable research direction. 
This will aid in intuitively identifying the sources of generated content that do not comply with standards, and accordingly taking appropriate remedial actions.


\paragraph{Combination of black-box LLMs and FL.} 
Currently, when combining LLMs with FL, researchers typically use a white-box approach, where the model's structure and parameters are fully transparent. This allows for a deep understanding of how the models work and enables adjustments to meet FL requirements. However, some high-performance LLMs, e.g., GPT-4~\cite{achiam2023gpt}, operate as black-box API services in real-world applications~\cite{he2022cater,he2022protecting}, meaning users can't access the internal workings of the model directly but interact with it through an API.
To effectively combine black-box LLMs with FL, knowledge distillation can be employed~\cite{hinton2015distilling}. For example, pre-trained LLMs act as teacher models, guiding the training of student models within the FL system. The teacher model's output, obtained via API calls, serves as pseudo-labels for FL training data. Student models' predictions are then aligned with these pseudo-labels to distill knowledge effectively~\cite{sun2023fedbpt}.

%% file: body/3-2-analysis.tex
\subsection*{3.2 Integration of Sub-technologies in FL with LLMs}

\subsubsection*{3.2.1 Current Status}

The training requirements for LLMs demand an immense amount of data and computational resources. 
In FL, distributed computing~\cite{auyoung2004resource} and privacy-preserving computation~\cite{yang2021toward} are considered effective tools to meet these demands. 
Existing research integrates these key technologies covered by FL with LLMs, aiming to address the practical issues LLMs face (Tabel \ref{tab:fl2llms}).

\begin{table}[t]
    \caption{Overview of the current state of sub-technologies within FL and their integration with LLMs. We list the existing research on the combination of each sub-technology with LLMs and analyze the benefits they bring. Subsequently, we provide a brief summary of their methodologies.}
    \label{tab:fl2llms}
    \centering
    \small
    \resizebox{\linewidth}{!}{%
    \begin{tabular}{c|ccc}
    \toprule
    Sub-technologies & Advantages of integration & Summary of method & References \\
    \midrule
    \multirow{2}{*}{Distributed computing} &
    Assessing computing resources & Aggregating computational capacities from multiple sources & \cite{zeng2023distributed,huang2024fast,wu2023fast} \\
     & Enhancing LLMs task generalization capabilities & Aggregating proprietary data from multi-party devices & \cite{hong2023mecta} \\
    \midrule
    \multirow{2}{*}{Privacy-preserving computation} &
    Improving prompt generation ability & Utilizing proprietary specific data to generate targeted prompts & \cite{yao2024survey,guo2023promptfl,chen2023can,chen2023hide} \\
     & Assisting with the Selection of CoT prompt & Balancing the generality and personalization in the selection of CoT prompts. & ~\cite{xing2023fedlogic,duan2023privacy,wang2023survey} \\
    \bottomrule
    \end{tabular}
    }
\end{table}

\paragraph{Distributed computing.} In FL, distributed computing helps LLMs by combining computing and data resources. This eases the workload for individual users during training and inference and boosts LLMs' ability to handle different tasks by merging data from multiple parties.

\textit{Assessing computing resources.} 
Training LLMs demands significant computational power. For example, LLaMA needs 2048 NVIDIA A100 GPUs for 21 days~\cite{touvron2023llama}, GPT-3-1.3B requires 64 Tesla V100 GPUs for 7 days~\cite{brown2020language}, and FLM utilizes 192 NVIDIA A800 GPUs for 22 days~\cite{li2023flm}. Such costs are manageable mainly by big tech firms like Microsoft and Google, limiting LLMs' progress. FedML and others combine FL with LLMs to share computing resources among participants, easing the burden during training and inference stages~\cite{zeng2023distributed,huang2024fast,wu2023fast}.

\textit{Enhancing LLMs task generalization capabilities.} 
LLMs are mainly trained on vast centralized datasets, e.g., GPT-NeoX-20B on Pile~\cite{black2022gpt} and LLaMA on comprehensive data including other LLMs' datasets~\cite{touvron2023llama}. Yet, these datasets don't cover all real-world knowledge, hampering models' adaptability. To address this, recent research integrates data from various sources using FL's distributed data processing~\cite{hong2023mecta}, aiming to enhance models' generalization, including data from medium-scale infrastructures and individual mobile devices.

\paragraph{Privacy-preserving computation.} Prompts play a crucial role in helping LLMs process complex tasks~\cite{liu2023prompt}. Public dataset prompts tend to be repetitive, and privacy regulations limit the use of private data for prompt generation. To improve this, current studies merge FL's privacy-preserving tech with prompt design in LLMs. This enhances personalized prompt matching, better meeting specific requirements.

\textit{Improving prompt generation ability.} 
LLMs improve their ability to understand complex tasks through prompt engineering~\cite{liu2023prompt}. However, to address privacy concerns, prompt designs often rely on publicly available data. This approach, while protecting privacy, limits the potential of prompt engineering from two aspects~\cite{yu2023federated}. Firstly, public datasets may not have access to specific domains or individual private information, hindering optimization for specialized fields. Secondly, using public datasets can result in generic prompt templates, leading to repetitive or uninspired model responses. Recent research integrates FL's privacy-preserving features with prompt generation~\cite{yao2024survey,guo2023promptfl,chen2023can,chen2023hide}, allowing for optimized prompts tailored to specific domains, thus enabling better adaptation to particular needs.

\textit{Assisting with the Selection of CoT prompt.} 
CoT reasoning, a method for eliciting quick and accurate responses from LLMs, is gaining attention in research~\cite{kojima2022large}. However, choosing the best prompts poses a challenge. Currently, prompt selection relies on trial and error, where users adjust prompts based on LLM responses. To improve the explainability of CoT prompt selection and balance universality and personalization across domains while protecting privacy, recent studies combine FL with LLMs~\cite{xing2023fedlogic,duan2023privacy,wang2023survey}. They propose the FedLogic framework~\cite{xing2023fedlogic}, which tackles prompt selection as a rule selection problem based on fuzzy scores, using LLMs as rule generators.

\subsubsection*{3.2.2 Challenges and Future Directions}

\paragraph{Additional communication overhead.} Although FL applied to LLMs can ease computational burdens, it introduces extra communication expenses. Due to the large number of LLM parameters, communication time might surpass training time significantly. Real-world network instability could worsen this issue~\cite{lim2020federated}. Additionally, extensive communication can harm the environment by raising carbon emissions.
Therefore, efficient distributed learning algorithms are crucial. These algorithms must tackle communication and computational challenges during LLM training and deployment across devices with varying capabilities and network conditions. Presently, training acceleration strategies, e.g., Deepspeed~\cite{rasley2020deepspeed}, Megatron~\cite{shoeybi2019megatron}, and BMTrain~\cite{zeng2023openbmb}, speed up LLM training via data parallelism, model parallelism, and pipeline parallelism~\cite{huang2019gpipe}. 
Applying these strategies within an FL environment is relatively straightforward and can, to some extent, address the local computational issues associated with FedLLMs. 
However, these strategies do not fully resolve communication challenges. 
A more effective approach involves employing model pruning~\cite{sun2023simple} and compression~\cite{zhu2023survey} techniques to reduce the complexity and size of the model, thereby alleviating computational and communication burdens without sacrificing model performance. 
Additionally, extending parameter-efficient fine-tuning methods to FedLLMs is also an effective solution.

\paragraph{Model property rights issues.} Training LLMs relies on vast, domain-specific datasets, making resulting models commercially valuable intellectual property. Ensuring ownership of these models is crucial, especially in distributed training scenarios like FL, which involve multiple collaborating parties~\cite{kairouz2021advances}. This increases the risk of model leaks and intellectual property infringement.
To tackle these challenges, it's essential to develop theoretical and methodological frameworks in artificial intelligence to identify ownership misappropriation and illegal claims. Authentication technologies should provide robust intellectual property protection without compromising model performance. Model watermarking is a promising solution~\cite{tekgul2021waffle}. It allows collaborative model updating and training while safeguarding private data and signatures. Implementing model watermarking in LLMs could effectively address intellectual property challenges.

%% file: body/3-3-analysis.tex
\subsection*{3.3 Overall Integration of LLMs and FL}

%
In this section, we discuss the benefits of FedLLMs compared to separate technologies, review current FedLLM trends, and examine potential challenges. Finally, we share our insights and suggestions for addressing these challenges.

\subsubsection*{3.3.1 Current Status}

\begin{table}[t]
    \caption{Comparison of current FedLLM frameworks. Following~\cite{ye2024openfedllm}, we adopt the following notation definitions. PT: parameter-efficient fine-tuning, IT: instruction-tuning, VA: value alignment-tuning, $N_{agg}$: number of supported FL aggregation algorithms, $N_{data}$: number of training datasets, $N_{eva}$: number of evaluation metrics.}
    \label{tab:comparison}
    \centering
    \small
    \resizebox{0.6\linewidth}{!}{
    \begin{tabular}{c|cccccc}
    \toprule
    Framework Name & PT & IT & VA & $N_{agg}$ & $N_{data}$ & $N_{eva}$ \\
    \midrule
    FATE-LLM~\cite{fan2023fate} & \checkmark & \texttimes & \texttimes & 1 & 1 & 4 \\
    Shepherd~\cite{zhang2024towards} & \checkmark & \checkmark & \texttimes & 1 & 1 & 1 \\
    FederatedScope-LLM~\cite{kuang2023federatedscope} & \checkmark & \checkmark & \texttimes & 1 & 3 & 3 \\
    OpenFedLLM~\cite{ye2024openfedllm} & \checkmark & \checkmark & \checkmark & 7 & 8 & $30_{+}$ \\
    \bottomrule
    \end{tabular}
    }
\end{table}

Compared to simply combining LLMs and FL sub-technologies, FedLLMs present a comprehensive framework, marking a new direction in privacy-preserving language model development. Its key advantages include: i) ensuring privacy while effectively integrating high-quality data from multiple parties for superior model training~\cite{fan2023fate}, ii) providing solutions for general task adaptation and large model training in specific areas~\cite{fan2023fate}, and iii) facilitating lifelong learning~\cite{yu2022towards}.

Several recent studies have explored integrating FL with LLMs~\cite{fan2023fate,zhang2024towards,kuang2023federatedscope,ye2024openfedllm}. Among them, the earliest studies publish perspective articles about FedLLMs~\cite{chen2023federated}. FATE-LLM~\cite{fan2023fate} investigates fine-tuning strategies within FL to reduce communication overhead, incorporating efficient fine-tuning methods~\cite{zhao2023survey}. However, its application was limited to traditional classification tasks. Subsequent works, i.e., FederatedScope-LLMs~\cite{kuang2023federatedscope} and Shepherd~\cite{zhang2024towards} expand into federated instruction-tuning but lack diverse training datasets. OpenFedLLM~\cite{ye2024openfedllm} adds a federated alignment-tuning mechanism, enhancing LLMs training. It conducts empirical analysis across datasets and compares FL aggregation methods ~\cite{mcmahan2017communication,reddi2020adaptive}.
Please refer to Table~\ref{tab:comparison} for a detailed comparison.

\begin{figure}[htbp]%
\centering
\includegraphics[width=1.0\textwidth]{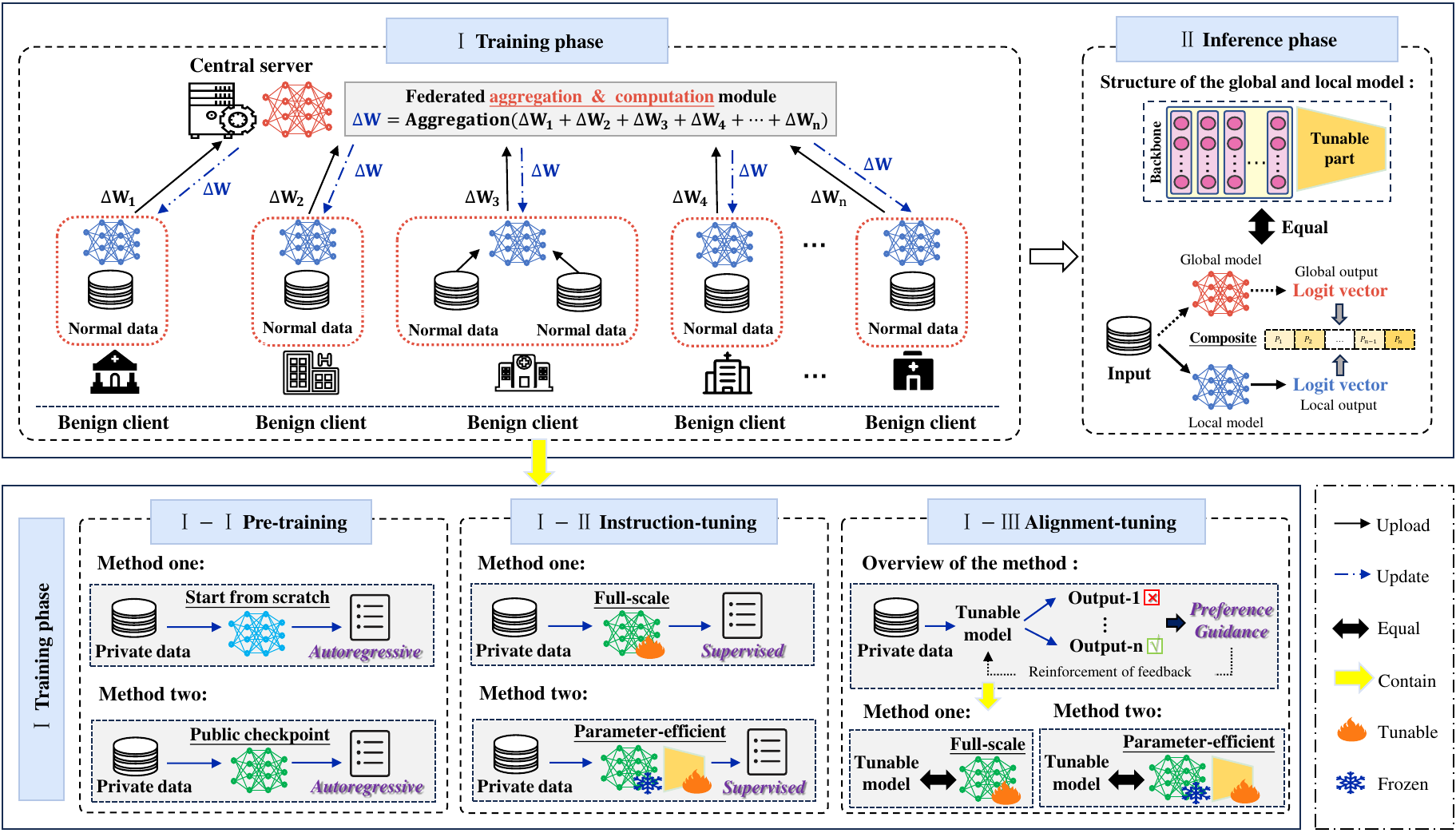}
\caption{The current implementation framework of FedLLMs. The use of FedLLM primarily includes two phases: training and inference. The training phase further comprises pre-training, instruction-tuning, and alignment-tuning. Our subfigures below detail feasible methods for implementing each part.}\label{fig:fedllms}
\end{figure}

Although existing FedLLMs frameworks differ in design and implementation, they follow the same core design principle: extending the training process of LLMs within an FL system. 
To construct a robust FedLLMs framework, we provide a detailed overview of the architecture of FedLLMs based on the aforementioned principle.
Refer to Figure \ref{fig:fedllms}, we divide the architectural structure of FedLLMs into two main phases: training and inference. 
In the training phase, the framework is further subdivided into pre-training, instruction-tuning, and alignment-tuning stages, with each stage offering a display of various specific implementation methods in the respective subfigures. 
During the inference phase, we expand the existing FedLLMs framework to allow clients to combine the outputs of local and global models when executing inferences, thereby enhancing the accuracy and adaptability of inference.

\subsubsection*{3.3.2 Challenges and Future Directions}

\paragraph{Synchronisation problems due to differences in computing resources.} 
In FL, exchanging gradient information between devices incurs high communication costs, especially when resources vary among participants. This issue is more pronounced in FedLLMs due to their larger parameter scales~\cite{bai2024federated}. Limited network bandwidth exacerbates the problem, potentially causing the dropout of some members and prolonging communication time~\cite{li2024snapfusion}.
To mitigate this issue, a naive strategy is collaborative computing~\cite{chang2017collaborative}, which limits the computational potential of resource-rich clients to match the capabilities of weaker clients. 
However, this approach can lead to a significant waste of computational resources. 
A more rational approach is to implement hierarchical aggregation~\cite{wang2021resource} within FedLLMs, where all clients do not communicate directly with the central server, but instead perform preliminary data aggregation within local groups (or clusters).
Each cluster is led by one or more clients with superior computational capabilities, responsible for collecting and aggregating data within the cluster before exchanging information with the central server or leaders of other clusters.

\paragraph{Incentive mechanisms for FedLLMs.} 
In the implementation of FedLLMs, creating a fair and effective incentive mechanism is crucial to encourage broader participation and collaboration among contributors, given the varying data volumes and computational capabilities among participants~\cite{zhan2020learning}. This involves balancing data contributions and computational resources, and aligning incentive strategies with each participant's investment and derived value.
Although incentive mechanisms for smaller models in traditional FL environments have been explored~\cite{zhan2021survey}, they are not directly applicable to LLMs due to their massive parameter size. Therefore, FedLLMs require specifically tailored incentive solutions.
Reinforcement learning has been highlighted as effective in designing FL incentive mechanisms, especially considering the characteristics of LLMs and their complementarity with reinforcement learning \cite{ouyang2022training}. Applying RL to FedLLMs and developing specialized incentive mechanisms for them seems to be a promising path forward.

\paragraph{Continuous data stream.} During FedLLM training, data streams continuously, unlike centralized LLMs where data comes in fixed batches~\cite{kirkpatrick2017overcoming,glorot2011domain}. Clients join and leave, potentially with data distributions different from the global model. Given LLMs' large size, training from scratch is resource-intensive. Integrating new data with minimal resources is a challenge.
Federated continual learning~\cite{criado2022non} is considered a viable solution. 
However, most existing methods are parameter-based~\cite{yoon2021federated}, which can be resource-intensive and inefficient when dealing with large-scale models. 
In contrast, loss-based approaches may be more suitable for FedLLMs. 
Furthermore, exploring the implementation of LLM model editing~\cite{mitchell2022memory} techniques in an FL environment, where only a subset of parameters is updated with each model iteration, could also help address this issue.

\paragraph{Personalisation issues in FedLLMs.} 
Current FedLLM frameworks prioritize training a unified model collaboratively, often overlooking personalization concerns. In the LLM domain, there are two main viewpoints: one advocates for universal LLMs with larger parameters~\cite{achiam2023gpt,chen2021evaluating}, while the other focuses on smaller, proprietary models for practical contexts like mobile devices. Considering personalized needs in FedLLMs is essential. 
A straightforward solution is to integrate existing federated personalization methods into the FedLLM framework for enhancement, such as model-agnostic personalization~\cite{fallah2020personalized}, hierarchical personalization~\cite{wu2021hierarchical}, and cluster-based federated learning~\cite{sattler2020clustered}.

\paragraph{Additional security and privacy threats.} 

LLMs combined with FL could worsen security and privacy risks, creating new challenges. Existing FedLLM frameworks often overlook these issues. To the best of our knowledge, we are the first to analyze security and privacy threats in FedLLMs. 
According to Usynin et al.~\cite{usynin2021adversarial}, we classify threats into utility-focused attacks and privacy-focused attacks (Table~\ref{tab:threats}). The former aims to impair model effectiveness, termed security threats, while the latter intends to compromise data privacy, termed privacy threats. We analyze these threats and their new variants in FedLLMs.

\begin{table}[t]
    \caption{Overview of attacks against FedLLMs systems. Based on the work by Usynin et al.~\cite{usynin2021adversarial}, we categorize the potential security and privacy threats faced by FedLLMs. Additionally, we highlight the common methods currently employed to defend against these threats. The notation definitions corresponding to these defenses are as follows. \small\rmnum{1}\normalsize: data analysis, \small\rmnum{2}\normalsize: update analysis, \small\rmnum{3}\normalsize: robust aggregation, \small\rmnum{4}\normalsize: model pruning, \small\rmnum{5}\normalsize: adversarial training, \small\rmnum{6}\normalsize: DP, \small\rmnum{7}\normalsize: knowledge distillation.}
    \label{tab:threats}
    \resizebox{\linewidth}{!}{
    \begin{tabular}{llll}
    \toprule
    Attack type & Attack goal & {Proposed defenses} & References \\
    \midrule
    \textbf{Security threats}  &   &   & \\
    Untargeted poisoning  &  Degrading utility of the target model  &  \rmnum{1},\rmnum{2},\rmnum{3},\rmnum{5} & \cite{tolpegin2020data,cao2019understanding,wallace2021concealed} \\
    Backdoor  &  Running an auxiliary learning task  &  \rmnum{1},\rmnum{3},\rmnum{4}  & \cite{gu2019badnets,chen2022badpre,kurita2020weight,shen2021backdoor} \\
    \textbf{Privacy threats}  &   &   & \\
    Membership inference  &  Inferring the presence of an individual record  &  \rmnum{6},\rmnum{7}  &  \cite{shokri2017membership,jagannatha2021membership,perez2022ignore,fowl2023decepticons}\\
    Attribute inference  &  Inferring sensitive value of a record  &  \rmnum{6} &  \cite{hayet2022invernet,mahloujifar2021membership,song2020overlearning}\\
    Model inversion  &  Reconstruction of training data  &  \rmnum{6},\rmnum{7}  &   \cite{fredrikson2015model,pan2020privacy}\\
    \bottomrule
    \end{tabular}
    }
\end{table}

\textit{Security threats.} Adversaries take a different approach, aiming to alter the learning protocol or undermine the model's utility~\cite{usynin2021adversarial}. In FedLLMs, the main threat to model performance is poisoning attacks. These attacks can be divided into untargeted and targeted (backdoor) poisoning attacks based on the attackers' goals~\cite{huang2011adversarial}.

\begin{itemize}[leftmargin=*]\setlength{\itemsep}{-\itemsep}
    \item Untargeted poisoning attacks. This attack involves minor manipulations of training data, where malicious actors introduce altered or distorted data samples into the federated dataset~\cite{cao2019understanding,tolpegin2020data}.
    This intentional bias or misguidance aims to disrupt the subsequent model training process. 
    %
    In FedLLMs, we incorporate textual data. 
    Introducing harmful noise to textual data, including tag inclusion, modification, or omission, is relatively easy to execute~\cite{wallace2021concealed}.
    %
    %
    %
    While typically used in image data, recent studies suggest optimized methods for perturbations on discrete data, i.e., textual data, expanding poisoning possibilities~\cite{carlini2023aligned,fang2023modeling}. In FL, a client could cause harm by sending corrupted updates, making FedLLMs vulnerable to adversarial perturbations~\cite{rodriguez2023survey}. Numerous studies suggest that FedLLMs are susceptible to poisoning attacks~\cite{schuster2021you,wan2022you,sun2023backdooring}, raising concerns about detection difficulty.

    \item Backdoor Attacks. This attack covertly manipulates models to exhibit normal behaviors but can be triggered by specific inputs to produce the adversary's desired output~\cite{gu2019badnets}.
    Unlike untargeted poisoning attacks, backdoor attacks involve the insertion or modification of precise input patterns~\cite{shen2021backdoor,chen2022badpre,dong2023investigating,mei2023notable}. 
    Backdoor attacks have extended in new ways within FedLLMs.
    During the instruction fine-tuning stage, LLMs are vulnerable to backdoor attacks. 
    %
    %
    Recent studies have acknowledged this risk, emphasizing the potential pathways for attackers to insinuate malicious commands~\cite{xu2021detoxifying,shu2023exploitability} and the concept of untrained vocabulary backdoor attacks on language models~\cite{huang2023training}. 
    Moreover, some studies perceive prompt injection attacks~\cite{wan2023poisoning,liu2023prompt} as a unique spin on backdoor attacks, with the compliance capabilities of LLMs being the primary target~\cite{shu2023exploitability}. 
    %
    Apart from novel types of backdoor attacks, the escalating complexity of models in FL settings fosters backdoor insertions. 
    This is attributable to the capacity of over-parameterized models to learn trigger features, even amidst label noise during training. 
    Furthermore, as models within FL are collectively utilized and contributed to by multiple clients, the scope for attacks and origins of backdoors inevitably broaden~\cite{sun2019can,bagdasaryan2020backdoor}, further increasing the risk of backdoor attacks.
\end{itemize}

\textit{Privacy threats.} 
This attack aims to access a client's private information. Due to the large number of parameters in LLMs, model extraction attacks are very costly. Therefore, this paper focuses on Membership Inference Attacks (MIA), attribute inference attacks, and model inversion attacks, and will not discuss model extraction attacks.

\begin{itemize}[leftmargin=*]\setlength{\itemsep}{-\itemsep}
    \item Membership Inference Attacks. This attack aims to predict whether a given data record is a member of the training dataset~\cite{shokri2017membership}.
    Given the fact that LLMs memorize training data~\cite{feldman2020does,carlini2021extracting,brown2021memorization,carlini2023quantifying}, the risk of MIA increases, especially if the memorized information includes personal or sensitive data~\cite{jagannatha2021membership,perez2022ignore}. 
    Hence, it is necessary to explore defense mechanisms against membership inference attacks in the FL environment.
    The integration of FL and LLMs could also potentially bring about novel manners of inference attacks.
    Recent research illustrates that the memory facet of LLMs significantly increases their susceptibility to privacy violations within the FL framework~\cite{gupta2022recovering}.
    %
    If the server is hypothesized to be dishonest or compromised, the structure of LLMs is prone to inference attacks~\cite{fowl2023decepticons}.
    %
    %
    How to defend against this threat is a question that needs to be explored in the future.
    
    \item Attribute inference attacks. This attack aims to recover characteristics of the training data learned by the model~\cite{song2020overlearning,gong2018attribute}.
    Due to LLMs commonly handling extensive textual data, attribute inference attacks carried out on text data are generally accomplished by obtaining the embedding vectors of text samples, thereby gaining access to the confidential attributes embedded within these samples.
    Prior research in this field has focused on the adjustment of these embeddings to apprehend semantic inter-relations existing between words, thereby predicting confidential data inherent in language models~\cite{hayet2022invernet}. 
    %
    %
    Engaging in FL increases the vulnerability to potential attacks, necessitating consideration of a broader range of attackers.
    %
    %
    Concurrently, white-box adversaries stand to benefit from unencrypted model updates, with a particular emphasis on gradient data, thereby delivering them a competitive edge in the process.

    \item Model inversion attacks. In attribute inference, it is crucial to acknowledge that attackers are typically required to hold additional information regarding their potential victims, such as personal identifiers (e.g., age and race), to exploit the association between this information and the sensitive features. 
    On the other hand, model inversion attacks primarily aim to reverse-engineer the internal representation produced by the model to reveal training data~\cite{fredrikson2014privacy,fredrikson2015model,he2019model}.
    In LLMs, this is typically done on the embedding rather than directly on the output~\cite{pan2020privacy,song2020information}.
    Recent advanced research suggests exploring embedding inversion attacks, showing that these attacks pose a higher privacy risk than attribute inference attacks~\cite{gu2023towards,morris2023text}. 
    Considering the FL scenario, the gradient leakage issue exacerbates this threat. 
    Studies demonstrate that the embeddings can be reconstructed based on leaked gradients~\cite{balunovic2022lamp,gupta2022recovering,chu2022panning}.
\end{itemize}

\paragraph{Outdated defense techniques.} Concerning the security and privacy threats discussed, existing defense strategies may not seamlessly apply to FedLLMs. We typically classify defense mechanisms into security defenses and privacy defenses (Table~\ref{tab:defenses}). By analyzing how current defenses might face challenges when applied to FedLLMs, we offer our perspectives and insights.

\begin{table}[t]
    \caption{Overview of defense techniques in FedLLMs systems. We enumerate the defense techniques currently in widespread use and have identified the types of attacks these defense methods can address in FedLLMs. The definitions of the symbols are as follows. \rmnum{1}: untargeted poisoning attack, \rmnum{2}: backdoor attack, \rmnum{3}: MIA, \rmnum{4}: attribute inference attack, \rmnum{5}: inversion attack.}
    \label{tab:defenses}
    \resizebox{\linewidth}{!}{
    \begin{tabular}{llll}
    \toprule
    Mitigation type & Summarize & {Mitigatable attacks} &  References \\
    \midrule
    \textbf{Security defences} &  &  & \\
    Data analysis & Analyzing data from other clients & \rmnum{1},\rmnum{2} &  \cite{cretu2008casting}\\
    Update analysis & Analyzing updates from various contributors & \rmnum{1} &   \cite{shen2016auror,andreina2021baffle}\\
    Robust aggregation & Replacing update averaging with aggregation & \rmnum{1},\rmnum{2} &  \cite{yin2018byzantine,wu2020federated,pillutla2022robust,blanchard2017machine}\\ 
    Model prunning & Dropping specific neurons/units of the model & \rmnum{2} &  \cite{dhillon2018stochastic,wu2020mitigating,sun2023simple}\\
    Adversarial training & Training the model on adversarial examples & \rmnum{1} &  \cite{ganin2016domain}\\
    \textbf{Privacy defences} & & & \\
    Differential privacy & Implementing targeted disturbances for the protocol &  \rmnum{3},\rmnum{4},\rmnum{5} &   \cite{dwork2014algorithmic,dagan2020pac,xu2023training,ozdayi2023controlling}\\
    Knowledge distillation  &  Transferring knowledge from the teacher model to the student model & \rmnum{3},\rmnum{5} & \cite{hinton2015distilling}\\
    \bottomrule
    \end{tabular}
    }
\end{table}

\textit{Security defenses.} Security defenses aim to alleviate the adverse effects of attacks on model performance. 
Our study mainly contemplates widely accepted methods, encompassing data analysis, update analysis, robust aggregation, model pruning, and adversarial training.

\begin{itemize}[leftmargin=*]\setlength{\itemsep}{-\itemsep}
    \item Data and update analysis. Data analysis involves evaluating data from other clients and implementing subsequent preprocessing~\cite{cretu2008casting}. 
    However, it is often impractical to apply under privacy-preserving conditions due to the need for access to user-specific local data~\cite{kairouz2021advances}.
    In contrast, update analysis reviews parameters from other clients to determine their necessity for aggregation. While effective, this technique requires access to client updates, potentially increasing the risk of privacy leakage. 
    The analysis often relies on outlier update analysis~\cite{shen2016auror,andreina2021baffle}, which is challenging due to the high dimensionality of LLMs model updates.
    Possible solutions may involve using dimensionality reduction techniques such as principal component analysis~\cite{tolpegin2020data} or implementing spectral anomaly detection with low-dimensional embeddings~\cite{li2020learning}.

    \item Robust aggregation. The aim of robust aggregation is to mitigate the negative impact of adversaries on the final model~\cite{blanchard2017machine,yin2018byzantine,guerraoui2018hidden,wu2020federated,pillutla2022robust}. Despite their significant potential, implementing these methods in deep learning models presents a major challenge~\cite{blanchard2017machine,xie2019zeno}. 
    %
    At the same time, it is important to consider the potential unintended consequences of modifying the aggregation mechanism, as this could have a negative impact on LLMs architectures.
    %
    Existing research has shown that FedAvg can negatively influence the attention mechanism of LLMs~\cite{ashraf2024transfed}. 
    %
    Thus, maintaining compatibility with LLMs architecture during this process is crucial.
    
    \item Model pruning. Model pruning assumes that the model's most weights contain knowledge relevant to the original task, while only a small portion is affected by poisoning attacks~\cite{dhillon2018stochastic}. This assumption suggests post-training defensive measures involving pruning the globally trained model to strengthen it against potential training-based attacks~\cite{wu2020mitigating}. 
    %
    %
    For deep network architectures like LLMs, specialized adaptations of model pruning techniques can be developed~\cite{grachev2019compression}, and exploration can also be carried out on mainstream LLMs~\cite{sun2023simple}.
    Ultimately, these pruning methods can be applied in the federated learning environment, enhancing overall robustness.

    \item Adversarial training. This approach trains the model using additional adversarial samples~\cite{ganin2016domain} to enhance the model's adversarial robustness against attacks. 
    However, there are potentially two issues with this approach when applied to FedLLMs. Firstly, generating adversarial examples for discrete data types such as text can be considerably more complex than for images, as direct perturbations in the embedding space may lead to significant semantic deviations due to minor disturbances~\cite{shayegani2023survey}. Secondly, the resource expenditure for generating adversarial samples in deep neural networks is substantial when using gradient-based adversarial perturbation methods like PGD~\cite{madry2017towards}.
    To address these challenges, a viable approach is to introduce an additional LLM that employs prompt engineering techniques to generate semantically similar adversarial samples\cite{mattern2023membership}. Alternatively, adversarial samples can also be generated through discrete optimization methods~\cite{tsymboi2023layerwise}. To further reduce overhead, updates can be selectively applied based on the importance of model parameters~\cite{kim2023roast}.
\end{itemize}

\textit{Privacy defenses.} Common privacy protection techniques include DP, knowledge distillation, regularization, and model pruning. However, the privacy benefits of regularization are limited, and certain techniques have been effectively bypassed, so we are not focusing on this method in this paper. Model pruning, which is a combined defense mechanism, has been found useful for privacy protection~\cite{wang2021against, usynin2021adversarial}. However, its use in privacy-sensitive situations may raise concerns as it could unintentionally reveal sensitive features of the training data. Therefore, this study categorizes it as a performance-focused defense. The main privacy-focused defense measures considered in this study are knowledge distillation and DP.

\begin{itemize}[leftmargin=*]\setlength{\itemsep}{-\itemsep}
    \item Knowledge distillation. Knowledge distillation allows the knowledge of a model to be transferred to a simpler model~\cite{hinton2015distilling}.
    Originally conceived to mitigate overfitting, knowledge distillation has evolved, with current research ingeniously combining it with DP principles~\cite{papernot2018scalable,fay2020decentralized}. 
    This innovative approach has given rise to new systems like private aggregation of teacher ensembles~\cite{papernot2017semi}. 
    These systems harness publicly available datasets to enable the transfer of knowledge from locally trained models to centralized models operating under DP mechanisms.
    In the federated setting, federated distillation is already a mature framework~\cite{jeong2018communication}. 
    Transferring the distillation technique of LLMs~\cite{gu2023knowledge} to FL is a direction worth exploring.

    \item Differential privacy. DP currently stands as the principal paradigm for privacy protection. 
    DP methods on language models include gradient perturbation-based approaches and embedding vector perturbation-based approaches~\cite{hu2023differentially}.
    The former adds noise to the gradients during network training, while the latter perturbs the word embeddings, aiming to protect privacy at the sample level (i.e., words or sentences). 
    However, in FedLLMs, privacy protection extends beyond the sample level to the user privacy level, aiming to safeguard each user's historical data. 
    Additionally, since only gradients are exchanged between clients in FedLLMs, methods based on embedding vector perturbation cannot be directly extended to FedLLMs.
    For gradient perturbation-based methods, although existing research provides theoretical privacy guarantees, two significant issues arise as the model scales up, and these issues are exacerbated in an FL environment. 
    Firstly, the computational and storage overhead of managing gradients increases~\cite{yu2021large}. 
    Secondly, the scale of noise required also increases~\cite{yu2021large}, which can adversely affect model performance to some extent.
    To address these challenges, a straightforward extension of the improved DP-SGD optimizer~\cite{yu2021large,li2023privacy,bu2022differentially,gupta2023jointly} to the FL environment is a viable direction. 
    Additionally, relaxing the level of DP to protect only the sensitive parts of samples using SDP-SGD~\cite{shi2022just} is another potential approach. 
    Finally, exploring the combination of DP with existing efficient parameter fine-tuning methods could also be a feasible strategy\cite{xu2023training,du2023dp}. 
    %
    %
    %
    %
\end{itemize}

%% file: body/4-discussion.tex
\section*{4. Discussion of Application for Combining LLMs and FL}

The integration of LLMs with FL promises to complement the advantages of both and effectively address their respective limitations. 
The resulting synergistic effect suggests that the amalgamation of LLMs with FL could be widely applied to various practical scenarios to tackle specific problems in the real world. 
Given that current research has explored the potential of the fusion of LLMs' sub-technologies with FL, as well as the integration of FL's sub-technologies with LLMs in application scenarios~\cite{zhuang2023foundation}, this paper will focus on discussing the feasible applications of FedLLMs in practice.

In light of the inherent distinguished characteristics of FedLLMs, we explore the broad range of application fields for FedLLMs. 
These applications mainly include healthcare, finance, education, and son on. 
Within these scenarios, deploying FedLLMs has the potential to address real industry problems, optimize service processes, and enhance overall efficiency and effectiveness.
Additionally, we also focus on the unique challenges faced by FedLLMs in these scenarios and provide an analysis of them.

\paragraph{Healthcare.} The healthcare scenario is one of the application areas that is intimately related to human well-being and is of great importance. 
Since the introduction of ChatGPT and other LLMs, numerous studies have applied these technologies in the healthcare field~\cite{singhal2023large,singhal2023towards,yang2024zhongjing}. 
It has been proven that LLMs have the capability to handle a variety of healthcare tasks, including but not limited to healthcare consultation recommendations~\cite{nov2023putting}, simplification of healthcare reports~\cite{jeblick2023chatgpt}, mental health analysis~\cite{yang2023evaluations}, and extraction of biohealthcare information~\cite{tang2023does}. 
To further tap into the potential of large models, recent research focuses on large models specially designed for the healthcare field, such as the Med-PaLM model~\cite{suzgun2022challenging,singhal2023towards}. 
In the United States Healthcare Licensing Examination, this model demonstrates performance comparable to professionals and gained broader recognition from the healthcare community in answering consumer health questions.
However, there is a risk of privacy breaches when the current LLMs upload patient health information to commercial servers that support model training~\cite{tang2023does,jeblick2023chatgpt}. 
This issue urgently needs to be addressed through technical means. 
FedLLMs offer an effective way to help healthcare institutions aggregate data from multiple parties to train their own healthcare-specific large models, tackling the aforementioned privacy challenges.

Although FedLLMs exhibit high potential for application in the healthcare scenario, the implementation of this technology is still limited by the unique characteristics of healthcare data and its strict usage regulations. 
The specific challenges faced include:
\begin{itemize}
    \item Data heterogeneity. Healthcare data often originates from various sources, including electronic health records, healthcare imaging, and laboratory results. 
    These data vary significantly in format, quality, and level of detail. 
    In FedLLMs, due to the involvement of multiple different participants, the types of data held by each party may also be inconsistent, further exacerbating the problem of data heterogeneity.
    \item Data incompleteness and imbalance. Healthcare units participating in FedLLMs often face issues of missing data or incomplete records, especially in scenarios involving long-term monitoring of patients.
    Additionally, data samples for certain diseases may be much less than for others, leading to data imbalance during model training, which can affect the model's generalization ability and accuracy.
    \item Model interpretability. The healthcare field has higher requirements for model interpretability compared to other sectors. 
    Healthcare decisions directly affect people's health, and doctors and patients usually need to clearly understand the basis of model predictions. 
    How to reflect the interpretability of models within the FedLLMs framework is an urgent issue to be addressed.
\end{itemize}

\paragraph{Finance.} The field of finance is one of the key areas where LLMs demonstrate their vast application potential. 
LLMs have been employed in a variety of financial tasks, including but not limited to financial reasoning~\cite{son2023beyond}, digital claims detection~\cite{shah2023zero}, financial named entity recognition~\cite{alvarado2015domain}, and financial sentiment analysis~\cite{araci2019finbert}. 
While general-purpose LLMs like ChatGPT have notable performances in the financial industry, they still cannot match the level of large models that are specifically trained and fine-tuned for the financial scenario, such as BloombergGPT~\cite{wu2023bloomberggpt}, FinGPT~\cite{yang2023fingpt}, etc.
However, LLMs tailored for the financial scenario require access to vast amounts of high-quality financial data~\cite{wu2023bloomberggpt}, which may exceed the capacities of some organizations. 
FedLLMs offer an innovative path for cultivating financial scenario-specific large models. 
Moreover, given that content generated by financial models could have significant impacts on markets, stringent alignment and adjustment of financial models is an indispensable step. 
The collaborative mechanism of FedLLMs can meet more complex and stricter alignment requirements, ensuring that aligned models adequately regard and reflect the interests of the majority of participants.

While FedLLMs introduce unprecedented new opportunities in the financial scenario, they also bring a series of new challenges:
\begin{itemize}
    \item High dynamism. Data in financial markets is highly dynamic and changes rapidly. 
    For instance, stock prices and interest rates can undergo significant changes within very short periods. 
    This requires the FedLLMs framework to support participants in rapidly updating language models in a short time, rather than relying on periodic retraining.
    \item High accuracy and reliability. Financial decisions often have significant financial implications, thus the information provided must be extremely accurate and reliable. 
    This poses higher accuracy standards for the inference process of the FedLLMs framework.
    \item Enhanced contextual understanding. Financial question-answering scenarios often involve complex contexts and multi-step logical reasoning. 
    When applying the FedLLMs framework for inference, it needs to possess strong contextual understanding capabilities, capable of handling coherent dialogue, remembering previous communications, and understanding complex query intentions.
\end{itemize}

\paragraph{Education.} The education scenario is also a key application area significantly influenced by LLMs. 
Recently, several pioneering research papers have explored the diverse applications of LLMs in educational settings~\cite{tan2023towards,kamalov2023new}, including teacher-student interactive collaboration, personalized learning experiences, and the automation of assessment processes. 
However, the application of LLMs in education can also bring a range of practical issues, such as homework plagiarism, the intrinsic biases of AI-generated content, over-reliance on LLMs, and the inequity in accessing resources for non-English-speaking countries~\cite{kasneci2023chatgpt}.
Against this backdrop, FedLLMs offer a solution for cultivating fair LLMs. 
The increase in participating parties and the richness of training data contribute to reducing biases present in LLMs and expanding their adaptability to multilingual environments. 
Through FedLLMs, it is possible to achieve multi-dimensional data collaboration, driving the creation of equitable and inclusive educational LLMs that consider and balance the needs of different languages and cultural backgrounds.

In the application within the educational scenario, the FedLLMs framework also faces several new challenges:
\begin{itemize}
    \item Complexity of different educational stages and backgrounds. In the FedLLMs framework, participant entities serve student groups that vary significantly in age, learning abilities, and background knowledge. 
    Therefore, the framework needs to possess the capability to adapt to these differences, in order to provide customized learning recommendations and content.
    \item Diversity of educational goals. Educational objectives are not limited to improving academic performance but also include emotional development, social skills, and growth in other non-academic areas. 
    In this context, FedLLMs need to consider these multifaceted factors to assess and propose recommendations for the holistic development of students.
    \item Strong guidance capability. An ideal educational LLM should guide students gradually toward finding the correct answers. 
    In the FedLLMs framework, enhancing the model's CoT reasoning capabilities is a critical issue that requires focused attention.
    \item Higher alignment requirements. In the educational field, given the limited discernment abilities of students at different age levels, there are higher demands for the alignment of models trained via FedLLMs. 
    Furthermore, the model should also be capable of refusing unreasonable requests of students.
\end{itemize}

%% file: body/5-conclusion.tex
\section*{5. Conclusion and Future Work}

Creating high-performing and robust LLMs relies on having sufficient high-quality data, which is often difficult and costly to obtain. 
To address the issue of data scarcity, researchers have incorporated FL techniques into LLMs, enabling the pooling of data from multiple parties for training while ensuring privacy. Additionally, integrating LLMs into FL helps address some specific challenges faced by FL, as LLMs possess exceptional task generalization capabilities. Numerous studies have demonstrated the complementarity of LLMs and FL in these domains. These studies include investigation of non-language foundation models, due to their potential for straightforward extension to LLMs, provide a broader perspective for our research. Given this complementarity, the research field combining LLMs with FL demonstrates significant potential for development. 
In this regard, this paper explores this research area, proposing a framework to organize ongoing efforts. 
We analyze advantages, challenges, and future directions, including potential applications in healthcare, finance, and education. 
This review aims to guide the development of integration technologies between LLMs and FL, emphasizing the need for unified evaluation benchmarks and datasets in future research.